\documentclass[10pt,twocolumn,letterpaper]{article}

\usepackage{iccv}
\usepackage{times}
\usepackage{epsfig}
\usepackage{graphicx}
\usepackage{amsmath}
\usepackage{amssymb}
\usepackage{graphics}
\usepackage{booktabs}
\usepackage{multirow}
\usepackage{enumitem}
\usepackage{bm}
\usepackage[innercaption]{sidecap}
\usepackage{upquote}
\usepackage{textcomp}
\usepackage{pifont}
\usepackage{etoolbox}
\usepackage{caption}
\usepackage{subcaption}
\usepackage[utf8]{inputenc}
\usepackage{xspace}
\usepackage{enumitem}
\usepackage{natbib}
\usepackage{lipsum}
\usepackage{subcaption}
\usepackage{wasysym}
\usepackage[dvipsnames]{xcolor}

\setcitestyle{numbers,square,citesep={,}}
\DeclareGraphicsExtensions{.jpeg,.png,.pdf}
\def\ie{\emph{i.e.}\xspace}

\newcommand{\vbar}{\,|\,}

\newcommand{\timesnarrow}{{\mkern-2mu\times\mkern-2mu}}

\newcommand{\partitle}[1]{\noindent\textbf{#1}}
\newcommand{\ptspace}{\vspace*{5pt}}

\DeclareFontFamily{U}{mathb}{\hyphenchar\font45}
\DeclareFontShape{U}{mathb}{m}{n}{
<-6> mathb5
<6-7> mathb6
<7-8> mathb7
<8-9> mathb8
<9-10> mathb9
<10-12> mathb10
<12-> mathb12}{}
\DeclareSymbolFont{mathb}{U}{mathb}{m}{n}
\DeclareMathSymbol{\llcurly}{\mathrel}{mathb}{"CE}
\DeclareMathSymbol{\ggcurly}{\mathrel}{mathb}{"CF}
\definecolor{colornode01}{HTML}{4E79A7}
\definecolor{colornode04}{HTML}{76B7B2}
\definecolor{colornode05}{HTML}{59A14F}
\definecolor{colornode06}{HTML}{EDC948}
\definecolor{colornode08}{HTML}{FF9DA7}
\definecolor{colornode09}{HTML}{9C755F}
\definecolor{lightgray}{gray}{0.6}
\iccvfinalcopy

\hypersetup{pdfstartview={FitH}}
\hypersetup{pdfborder={0 0 0}}
\hypersetup{pdfpagemode={UseNone}}
\hypersetup{pdfcreator={Noureldien Hussein}}
\hypersetup{pdfproducer={Noureldien Hussein}}
\hypersetup{pdftitle={VideoGraph: Recognizing Minutes-Long Human Activities in Videos}}
\hypersetup{colorlinks=true}
\hypersetup{pdfsubject={Computer Vision}}
\hypersetup{pdfauthor={Noureldien Hussein, Efstratios Gavves, Arnold~W.M. Smeulders}}
\hypersetup{pdfkeywords={Computer Vision, Complex Actions, Action Recognition, VideoGraph, Human Activities, Long-range, Temporal Modeling}}

\begin{document}
\title{VideoGraph: Recognizing Minutes-Long Human Activities in Videos}
\author{
Noureldien~Hussein, Efstratios~Gavves, Arnold~W.M.~Smeulders
\\
QUVA~Lab, University~of~Amsterdam
\\
{\tt\small\{nhussein, egavves, a.w.m.smeulders\}@uva.nl}}
\maketitle
\begin{abstract}
Many human activities take minutes to unfold.
To represent them, related works opt for statistical pooling, which neglects the temporal structure.
Others opt for convolutional methods, as CNN and Non-Local. While successful in learning temporal concepts, they fall short of modeling minutes-long temporal dependencies.
We propose VideoGraph, a method to achieve the best of two worlds: represent minutes-long human activities and learn their underlying temporal structure.
To represent human activities, VideoGraph learns a soft version of an undirected graph.
The graph nodes are deterministic and are learned entirely from video datasets, making VideoGraph applicable to video understanding tasks without node-level annotation.
While the graph edges are probabilistic and are learned in a soft-assignment manner.
The result is improvements over related works on benchmarks: Breakfast, Epic-Kitchens and Charades.
Besides, we demonstrate that VideoGraph is able to learn the temporal structure of human activities in minutes-long videos.
\end{abstract}

\section{Introduction}\label{sec:intro}
Human activities in videos can take many minutes to unfold, each is packed with plentiful of fine-grained visual details.
Take for example two activities: ``making pancake" or ``preparing scrambled eggs".
The question is what makes a difference between these two activities?
Is it the fine-grained details in each, or the overall painted picture by each? Or both?

\begin{figure}[!t]
\begin{center}
\includegraphics[trim=4mm 4mm 2mm 0mm,width=1.0\linewidth]{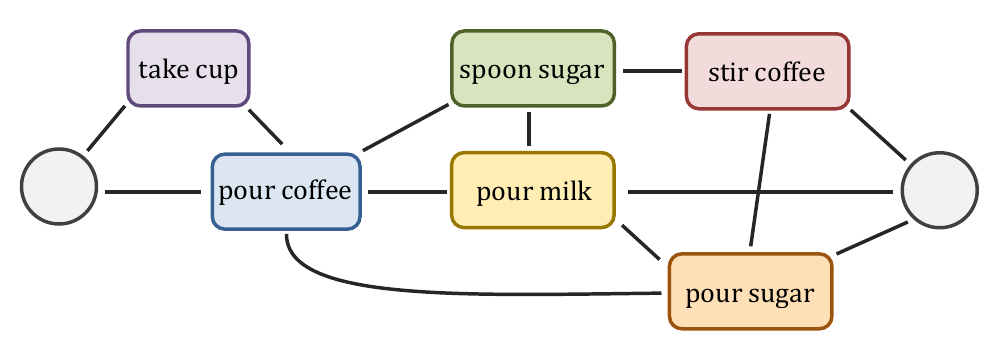}
\end{center}
\caption{The activity of ``preparing coffee" can be represented as undirected graph of unit-actions.
We are inspired by graphs to represent this activity.
The reason is that a graph can portray the many ways one can carry out such activity.
More over, it preserves the temporal structure of the unit-actions.
Reproduced from~\cite{kuehne2014language}.}
\label{fig:1-1}
\vspace*{-5mm}
\end{figure}

The goal of this paper is to recognize minutes-long human activities as defined by~\cite{kuehne2014language}, also referred to as complex actions in~\cite{hussein2018timeception}.
A long-range activity consists of a set of unit-actions~\cite{kuehne2014language}, also known as one-actions~\cite{hussein2018timeception}.
For example, the activity of ``making pancakes" includes unit-actions: ``cracking egg", ``pour milk" and ``fry pancake".
Some of these unit-actions are crucial to distinguish the activity. 
For example, the unit-action ``cracking egg" is all what needed to discriminate the activity of ``making pancakes" from ``preparing coffee".
Also, long-range activity is recognized only in its entirety, as its unit-actions are insufficient by themselves.
For example, only a short video snippet of unit-action ``cracking egg” cannot tell apart ``making  pancake" from ``preparing scrambled eggs", as both activities share the same unit-action ``cracking egg”.
Added to this, the temporal order of unit-actions for a specific activity may be permuted.
There exist different orders of how we can carry out an activity, like ``prepare coffee", see figure~\ref{fig:1-1}.
Nonetheless, there exist some sort of temporal structure for such activity.
One can start ``preparing coffee" by ``taking cup" and usually end up with ``pour sugar" and ``stir coffee".
So, to recognize long-range human activities, goals to be met are: modeling the temporal structure of the activity in its entirety, and occasionally paying attention to its fine-grained details.

There exist two distinct approaches for long-range temporal modelling. The first approach is orderless modeling.
Statistical pooling~\cite{hussein2017unified} and vector encoding~\cite{duta2017spatio, girdhar2017actionvlad} are used to aggregate video information over time.
The upside is the ability to address seemingly minutes- or even hours-long videos.
The downside, however, is the inability to learn temporal patterns and the arrow-of-time~\cite{ghodrati2018video}.
Both are proven to be crucial for some tasks~\cite{sigurdsson2017actions, huang2018makes}.
The second approach is order-ware modelling. 3D CNN is proven to be successful in learning spatiotemporal concepts for short video snippets with strict temporal pattern~\cite{carreira2017quo}.
Careful design choices enable them to model up to minute-long temporal dependencies~\cite{hussein2018timeception}.
But for minutes-long human activities, the strict temporal pattern no longer exists.
So, the question arises: how to model the temporal structure of minutes or even hour-long human activities?

This paper proposes VideoGraph, a graph-inspired representation to achieve the aforementioned goal.
A soft version of undirected graph in learned completely from the dataset.
The graph nodes represent the key latent concepts of which the human activity is composed.
These latent concepts are analogous to one-actions.
While the graph edges represent the temporal relationship between these latent concepts, \emph{i.e.} the graph nodes.
VideoGraph has the following novelties.
\textit{i.} In its graph-inspired representation, VideoGraph models human activity for up to thirty-minute videos, whereas the state-of-the-art is one minute~\cite{hussein2018timeception}.
\textit{ii.} A proposed node embedding block to learn the graph nodes from data.
This circumvents the node annotation burden for long-range videos, and makes VideoGraph extensible to video datasets without node-level annotation.
\textit{iii.} A novel graph embedding layer to learn the relationships between graph nodes.
The outcome is representing the temporal structure of long-range human activities.
The result is achieving improvements on benchmarks for human activities: Breakfast~\cite{kuehne2014language}, Epic-Kitchens~\cite{damen2018scaling} and Charades~\cite{sigurdsson2016hollywood}.

\section{Related Work}\label{sec:related_work}
\ptspace
\partitle{Orderless \textit{v.s.} Order-aware Temporal Modeling.}
Be it short-, mid-, or long-range human activities, when it comes to temporal modeling, related methods are divided into two main families: orderless and order-aware.
In orderless methods, the main focus is the statistical pooling of temporal signals in videos, without considering their temporal order or structure.
Different pooling strategies are used, as max and average pooling ~\cite{hussein2017unified}, attention pooling ~\cite{girdhar2017attentional}, and context gating ~\cite{miech2017learnable}, to name a few.
A similar approach is vector aggregation, for example: Fisher Vectors~\cite{oneata2013action} and VLAD ~\cite{duta2017spatio, girdhar2017actionvlad}.
Although statistical pooling can trivially scale up to extremely long sequences in theory, this comes at a cost of losing the temporal structure, reminiscent of Bag-of-Words losing spatial understanding.

In order-aware methods, the main attention is payed to learning structured or ordered temporal patterns in videos.
For example, LSTMs~\cite{li2017concurrent, donahue2015long}, CRF~\cite{sigurdsson2017asynchronous}, 3D CNNs~\cite{xu2017r, carreira2017quo, tran2018closer, xie2017rethinking, wang2017non}.
Others propose temporal modeling layers on top of backbone CNNs, as in Temporal-Segments~\cite{wang2016temporal}, Temporal-Relations~\cite{zhou2017temporal} and Rank-Pool ~\cite{fernando2017rank}.
The outcome of order-aware methods is substantial improvements over their orderless counterparts in standard benchmarks ~\cite{kay2017kinetics, kuehne2011hmdb, soomro2012ucf101}.
Nevertheless, both temporal footprint and computational cost remain the main bottlenecks to learn long-range temporal dependencies.
The best methods ~\cite{hussein2018timeception, wang2017non} can model as much as 1k frames ($\sim$30 seconds), which is a no match to minutes-long videos.
This paper strives for the best of two worlds: learning the temporal structure of human activities in minutes-long videos.

\ptspace
\partitle{Short-range Actions \textit{v.s.} Long-range Activities.}

Huge body of work is dedicated to recognizing human actions that take few seconds to unfold. Examples of well-established benchmarks are: Kinetics~\cite{kay2017kinetics}, Sports-1M~\cite{karpathy2014large}, YouTube-8M~\cite{abu2016youtube}, Moments in Time~\cite{monfort2018moments}, 20B-Something~\cite{goyal2017something} and AVA~\cite{gu2017ava}.
For these short- or mid-range actions, ~\cite{sigurdsson2017actions} demonstrates that a few frames suffice for a successful recognition.
Other strands of work shift their attention to human activities that take minutes or even an hour to unfold.
Cooking-related activities are good examples, as in YouCook~\cite{zhou2018towards}, Breakfast~\cite{kuehne2014language}, Epic-Kitchens~\cite{damen2018scaling}, MPII Cooking~\cite{rohrbach2016recognizing} or 50-Salads~\cite{stein2013combining}.
Other examples include instructional videos: Charades~\cite{sigurdsson2016hollywood}, and unscripted activities: EventNet~\cite{ye2015eventnet}, Multi-THUMOS~\cite{yeung2018every}.

In all cases, several works~\cite{kuehne2014language, hussein2018timeception, rohrbach2016recognizing, gaidon2011actom} define the differences between short- and long-range human actions, albeit with a different naming or terms.
We follow the same definition of~\cite{kuehne2014language}.
More formally, we use \textit{unit-actions} to refer to fine-grained, short-range human actions, and \textit{activities} to refer to long-range complex human activities.

\ptspace
\partitle{Graph-based Representation.}
Earlier, graph-based representation has been used in storytelling ~\cite{kim2014reconstructing,xiong2015storyline}, and video retrieval~\cite{pan2001videograph}.
Different works use graph convolutions to learn concepts and/or relationships from data ~\cite{niepert2016learning, defferrard2016convolutional,kipf2016semi}.
Recently, graph convolutions are applied to image understanding~\cite{yunpeng2018Graph}, video understanding~\cite{wang2018videos, Huang2018finding, girdhar2018video, huang2018neural} and question answering~\cite{wang2018make}.
Despite their success in learning structured representations from video datasets, the main limitation of graph convolution methods is requiring the graph nodes and/or edges to be known a priori.
Consequently, when node or frame-level annotations are not available, using these methods is hard.
In contrast, this paper aims for a graph-inspired representation in which the graph nodes are fully inferred from data.
The result is that our paper is extensible to datasets without node-level annotations.

\ptspace
\partitle{Self-Attention}
is used extensively in language understanding~\cite{lin2017structured}.
The recently proposed the transformer block shows substantial improvements in machine translation~\cite{vaswani2017attention}, image recognition~\cite{wang2017non} and video understanding~\cite{girdhar2018video, wu2018long} or even graph representations~\cite{velivckovic2017graph}.
The transformer block~\cite{wu2018long} attends to a local feature conditioned on both local and global context.
That is why it outperforms the self-attention mechanism~\cite{li2018videolstm, du2018interaction, yang2018action}, which is conditioned on only the local feature.

A video of human activity consists of short snippets of unit-actions.
This paper is inspired by all these attention mechanisms to attend to a unit-action (\ie local feature) based on the surrounding activity (\ie global context).

\begin{figure*}[!ht]
\begin{center}
\includegraphics[trim=0mm 4mm 0mm 5mm,width=0.9\linewidth]{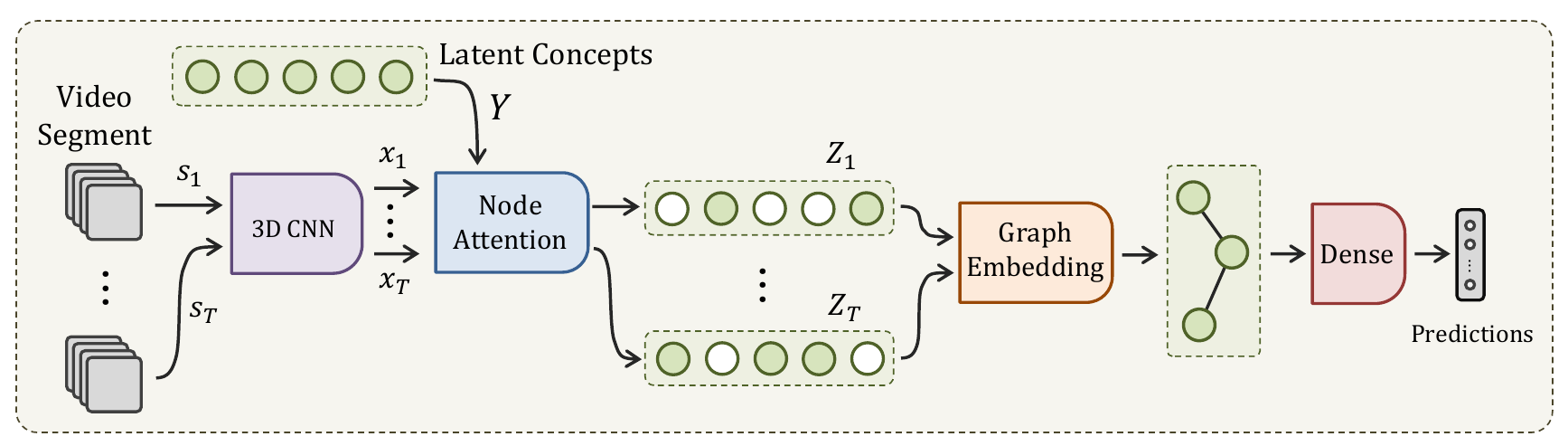}
\end{center}
\caption{
Overview of VideoGraph.
It takes as input a video segment $s_i$ of 8 frames from an activity video $v$. Then, it represents it using standard 3D CNN, \textit{.e.g} I3D.
The corresponding feature representation is $x_i$.
Then, a node attention block attends to a set of $N$ latent concepts based on their similarities with $x_i$, which results in the node-attenative representation $Z_i$.
A novel graph embedding layer then processes $Z_i$ to learn the relationships between its latent concepts, and arrives at the final video-level representation.
Finally, an MLP is used for classification.}
\label{fig:3-1}
\vspace*{-5mm}
\end{figure*}

\section{Method}\label{sec:method}
\partitle{Motivation.}
We observe that a minutes-long and complex human activity usually is sub-divided into unit-actions. Similar understanding is concluded by~\cite{kuehne2014language, hussein2018timeception}, see Fig.~\ref{fig:1-1}.
So, one can learn the temporal dependencies between these unit-actions using methods for sequence modeling in videos, as LSTM~\cite{li2017concurrent} or 3D CNN~\cite{xie2017rethinking}.
However, these methods face the following limitations.
First, such activities may take several minutes or even hours to unfold.
Second, as video instances of the same activity are usually wildly different, there is no single temporal sequence that these methods can learn.
For example, one can ``prepare coffee'' in many different ways, as the various paths in Fig.~\ref{fig:1-1} indicate.
Nevertheless, there seems to be an over-arching weak temporal structure of unit-actions when making a coffee.

We are inspired by graphs to represent the temporal structure of the human activities in videos.
The upside is the ability of a graph-based representation to span minutes- or even hour-long temporal sequence of unit-actions while preserving their temporal relationships.
The proposed method, VideoGraph, is depicted in Fig.~\ref{fig:3-1}, and in the following, we discuss its details.

\ptspace
\partitle{VideoGraph.}

We start from a video $v$ comprising $T$ randomly sampled video segments $v=\{s_i \vbar i = 1, 2, ..., T\}$.
Each segment $s_i$ is a burst of 8 successive video frames, and represented as feature $x_i \in \mathbb{R}^{1 \timesnarrow H \timesnarrow W \timesnarrow C}$ using standard 3D CNN, for example I3D~\cite{carreira2017quo}, where $C$ is the number of channels, $H, W$ are height and width of the channels.
Our goal is to construct an undirected graph $\mathcal{G}= (\mathcal{N},\mathcal{E})$ to represent the structure of human activity in video $v$.
The graph nodes $\mathcal{N}$ would then capture the key unit-actions in the activity.
And the graph edges $\mathcal{E}$ would capture the temporal relationship between these nodes (\textit{i.e.}  unit-actions).

\ptspace
\partitle{Learning The Graph Nodes.}
In a dataset of human activities, unit-actions can be thought of as the dominant {\em latent} short-range concepts.
That is, unit-actions are the building blocks of the human activity.
So, in a graph-inspired representation of the activity, these unit-actions can act as the graph nodes $\mathcal{N}$.
Assuming that it is prohibitively expensive to have unit-actions annotation for minutes-long videos, a challenge is how to represent them?
In other words, how to represent the graph nodes?
As a remedy, we opt for learning a set of $N$ latent features $Y$, $Y=\{y_j \vbar j = 1, 2, ..., N\}, Y \in \mathbb{R}^{N \timesnarrow C}$.
These features $Y$ then become the vector representation of the graph nodes $\mathcal{N}$, \textit{i.e.} $Y \equiv \mathcal{N}$.

\begin{figure}[!ht]
\begin{center}
\includegraphics[trim=1mm 5mm 1mm 0mm,width=1.0\linewidth]{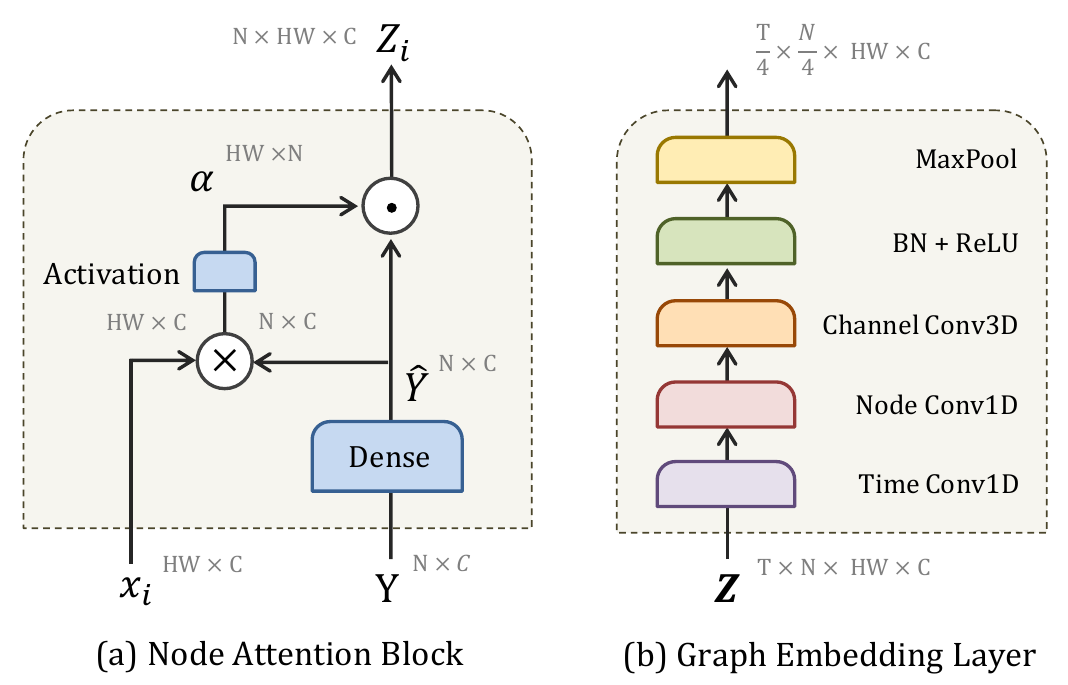}
\end{center}
\caption{(a) Node attention block measures similarities $\bm{\alpha}$ between segment feature $x_i$ and learned nodes $\hat{Y}$.
Then, it attends to each node in $\hat{Y}$ using $\bm{\alpha}$. The result is the node-attentive feature $Z_i$ expressing how similar each node to $x_i$.
(b) Graph Embedding layer models a set of $T$ successive node-attentive features $\mathbf{Z}$ using 3 types of convolutions.
\textit{i.} Timewise Conv1D learns the temporal transition between node-attentive features $\{Z_{i}, ..., Z_{i+t}\}$.
\textit{ii.} Nodewise Conv1D learns the relationships between nodes $\{ z_{i,j}, ..., z_{i,j+n} \}$.
\textit{iii.} Channelwise Conv3D updates the representation for each node $z_{ij}$.}
\label{fig:3-2}
\end{figure}

A problem, however, is how to correlate each video feature $x_i$ with each node in $Y$.
To solve this, we propose the node attention block, inspired by self-attention block~\cite{wang2017non, vaswani2017attention, girdhar2018video}, shown in Fig.~\ref{fig:3-2}{\color{red}a}.
The node attention block takes as an input a feature $x_i$ and all the node features $Y$.
Then, it transforms the initial representation of the nodes from $Y$ into $\hat{Y}$, using one hidden layer MLP with weight and bias $w \in \mathbb{R}^{C \timesnarrow C}, b \in \mathbb{R}^{1 \timesnarrow C}$.
This transformation makes the nodes learnable and better suited for the dataset inhand.
Then, a dot product $\otimes$ is used to measure the similarity between $x_i$ and $\hat{Y}$.
An activation function $\sigma$ is applied on the similarities to introduce non-linearity.
The result is the activation values $\bm{\alpha} \in \mathbb{R}^{H \timesnarrow W \timesnarrow N}$.
The last step is multiplying all the nodes $\hat{Y}$ with the activation values $\bm{\alpha}$, such that we attend to each node $\hat{y}_j$ by how much it is related to the feature $x_i$.
Thus, the node attention block outputs the attended nodes $Z_i = \{ z_{ij} \vbar j = 1, 2, ...,N\}, Z_i \in \mathbb{R}^{N\timesnarrow H \timesnarrow W \timesnarrow C}$. We refer to $Z_i$ as node-attentive feature, and we refer to $z_{ij}$ as the $j$-th node feature in $Z_i$. More formally,
\begin{align} 
\hat{Y} & = w * Y + b
\label{eqn:3-1}
\\ 
\bm{\alpha} & = \sigma (x_i * \hat{Y}^{T})
\label{eqn:3-2}
\\
Z_i & = \bm{\alpha} \odot \hat{Y}
\nonumber
\\
  & = \alpha_j \odot y_j, \;\; j = 1, 2, ..., N
\label{eqn:3-3}
\end{align}
Hence, the vector representation of all video segments is a 5D tensor $\mathbf{Z} = \{ Z_1, Z_2, ..., Z_T \}, \mathbf{Z} \in \mathbb{R}^{T \timesnarrow N \timesnarrow H \timesnarrow W \timesnarrow C}$. The names of 5 dimensions in $\mathbf{Z}$ are: timesteps, nodes, width, height and channels.
From now on, we use these 5 dimensions to express feature vectors and convolutional kernels.

In sum, the node attention block takes a feature $x_i$, corresponding to a short video segment $s_i$ and measures how similar $\bm{\alpha}$ it is to learned set of latent concepts $\hat{Y}$.
The similarities $\bm{\alpha}$ are then used to attend to the latent concepts.
This is crucial for recognizing long-range videos, where the network is not feed-forwarded only with a short video segment $x_i$ but with global representation $Y$.
This gives the network the ability for focus on both local video signal $x_i$ and global learned context $\hat{Y}$.

Our node attention block is different from the non-local counterpart~\cite{wang2017non} in twofold.
First, the attention values are conditioned on local $x_i$ and global $\hat{Y}$ signals.
Second, non-local does tensor product between attention values $\bm{\alpha}$ and local signal $x_i$, while we attend by scalar multiplication between $\bm{\alpha}, \hat{Y}$ to retrain the node dimension. Lastly, our node attention block is much more simpler than the non-local, as we use only one fully-connected layer.

\ptspace
\partitle{Learning The Graph Edges.}
Up till now, we have learned the graph nodes $\hat{Y}$.
We have also represented each video segment $s_i$ in terms of the nodes, as node-attentive feature $Z_i$.
Next, we would like to learn the graph edges $\mathcal{E}$, and arrive at the final graph structure.
To this end, we propose a novel graph embedding layer, shown in Fig.~\ref{fig:3-2}{\color{red}b}.
Regarding the graph edges, we are interested in two types of relationships.
First, we are interested in the relationship between graph nodes.
Loosely speaking, if nodes stand for unit-actions as ``pour milk", ``crack egg", we would like to learn how correlated are these two unit-actions when used in different activities as ``make pancake" or ``prepare coffee".
Second, we are interested in how the graph nodes transition over time.
For instance, we want to encode the significance of unit action ``pour milk" comes after or before ``crack egg" when it comes to recognizing ``make pancake".
Let's take $t$ successive video segments $\{ s_i, ..., s_{i+t} \}$.
When processed by CNN and node attention block, they are represented as $\{ Z_{i}, ..., Z_{i+t} \}$.
To learn the temporal transition between them, we apply a one-dimensional convolution, (Conv1D) on the temporal dimension only.
These timewise Conv1D, proposed by~\cite{hussein2018timeception}, are efficient in learning temporal concepts.
One kernel learned by timewise Conv1D is the 5D tensor $k^T \in \mathbb{R}^{t \timesnarrow 1 \timesnarrow 1 \timesnarrow 1 \timesnarrow 1}$, where $t$ is the kernel size.
In total, we learn $C$ kernels to keep the channel dimension of the features $\mathbf{Z}$ unchanged.

\begin{figure}[!ht]
\begin{center}
\includegraphics[trim=0mm 6mm 0mm 0mm,width=1.0\linewidth]{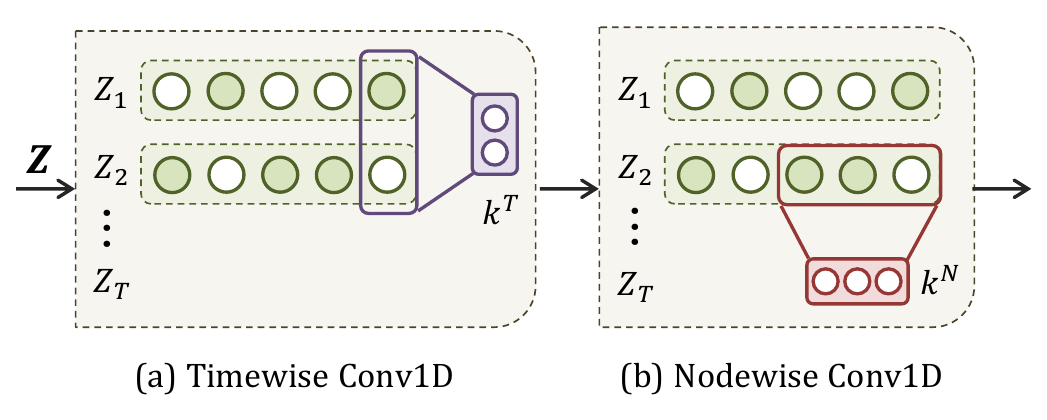}
\end{center}
\caption{(a) Timewise Conv1D learns the temporal transition between successive nodes-embeddings $\{ Z_{i}, ..., Z_{i+t} \}$ using kernel $k^T$ of kernel size $t$.
(b) Nodewise Conv1D learns the relationships between consecutive nodes $\{ z_{i,j}, ..., z_{i,j+n} \}$ using kernel $k^N$ of kernel size $n$.}
\label{fig:3-3}
\end{figure}

Besides learning the temporal transition between node-attentive features $\{Z_{i}, ..., Z_{i+t}\}$, we also want to learn the relationship between the nodes themselves $ \{ z_{ij} \vbar j = 1, 2, ..., N\}$ inside each node-attentive feature $Z_i$.
The problem is that the adjacency matrix, which defines the graph structure, is unkown.
A naive approach is to assume all nodes are connected.
This leads to an explosion of $N^2$ edges -- prohibitive to learn.
To overcome this, we restrict the number of adjacents (\textit{i.e.} neigbours) each node $z_{ij}$ can have.
In other words, we assume that each node $z_{ij}$ is adjacent to only $n$ other nodes.
This makes it possible to learn edge weights using one-dimensional convolution, applied on only the node dimension of $Z_i$.
We call this convolution nodewise Conv1D.
One kernel learned by nodewise Conv1D is the 5D tensor $k^N \in \mathbb{R}^{1 \timesnarrow n \timesnarrow 1 \timesnarrow 1 \timesnarrow 1}$, where $n$ is the kernel size.
In sum, we learn $C$ kernels to keep the channel dimension of the features $\mathbf{Z}$ unchanged.

Both timewise and nodewise Conv1D learn graph edges separately for each channel in the features $\mathbf{Z}$.
That is why we follow up with a typical spatial convolution (Conv2D) to model the cross-channel correlations in each node feature $z_{ij}$.
Spatial Conv2D learns $C$ different kernels, each is the 5D tensor $k^C \in \mathbb{R}^{1 \timesnarrow 1 \timesnarrow 1 \timesnarrow 1 \timesnarrow C}$.

Having learned the graph edges using convolutional operations, we proceed with BatchNormalization and ReLU non-linearity.
Finally, we downsample the entire graph representation $\mathbf{Z}$ over both time and node dimensions using MaxPooling operation.
It uses kernel size $3$ and stride $3$ for both the time and node dimensions.
Thus, after one layer of graph embedding, the result graph representation is reduced from $T \timesnarrow N \timesnarrow H \timesnarrow W \timesnarrow C$ to $ (T/3) \timesnarrow (N/3) \timesnarrow H \timesnarrow W \timesnarrow C$.

\section{Experiments}\label{sec:experiments}
\ptspace
\partitle{Implementation.}
When training VideoGraph on a video dataset, we uniformly sample $T=64$ video segments from each video $v$.
One segment $s_i$ is a burst of $8$ successive frames.
When the $64$ segments are fed-forward to I3D up to the last convolutional layer \texttt{res5\_c}, the corresponding convolutional features for the entire video is $\textbf{X} = \{ x_i \vbar i = 1, 2, ..., 64 \}, \textbf{X} \in \mathbb{R}^{64 \times 7 \timesnarrow7 \timesnarrow 1024}$.
We use $N = 128$ as the number of latent concepts.
Both the video-level features $\textbf{X}$ and latent concepts $Y \in \mathbb{R}^{128 \times 1024}$ are fed-forward to the node attention block.
The result is the graph representation $ \mathbf{Z} \in \mathbb{R}^{128 \timesnarrow 64 \times 7 \timesnarrow7 \timesnarrow 1024}$.
Then, $\mathbf{Z}$ is passed to graph embedding layers to learn node edges and reduce the feature representation.
In graph embedding layer, we use kernel size $t=7$ for the timewise Conv1D and kernel size $n=7$ for the nodewise Conv1D.
In total, we use 2 successive layers of graph embedding.
Their output feature is then feed-forwarded to a classifier to arrive at the vide-level predictions.
The classifier uses 2 fully-connected layers with BatchNormalization and ReLU non-linearity.
We use softmax as the final activation for single-label classification or sigmoid for multi-label classification.

VideoGraph is trained with batch-size 32 for 500 epoch.
It is optimized with SGD with 0.1, 0.9 and 0.00001 as learning rate, momentum and weight decay, respectively.
It is implemented using TensorFlow~\cite{tensorflow2015-whitepaper} and Keras~\cite{chollet2015keras}.

\subsection{Datasets}
As this paper focus on human activities spanning many minutes, we choose to conduct our experiments on the following benchmarks: Breakfast~\cite{kuehne2014language}, Epic-Kitchens~\cite{damen2018scaling} and Charades~\cite{sigurdsson2016hollywood}.
Other benchmarks for human activities contain short-range videos, \textit{i.e.} a minute or less, thus do not fall within the scope of this paper.

\ptspace
\partitle{Breakfast}
is a dataset for task-oriented human activities, with the focus on cooking.
It is a video classification task of 12 categories of breakfast activities.
It contains 1712 videos in total, 1357 for training and 335 for test.
The average length of videos is 2.3 minutes.
The activities are performed by 52 actors, 44 for training and 8 for test.
Having different actors for training and test splits is a realistic setup for testing generalization.
Each video is represents only one category of focus activity.
Besides, each video has temporal annotation of unit-actions comprising the activity.
In total, there are 48 classes of unit-actions.
In our experiments, we only use the activity annotation, and we don't use the temporal annotation of unit-actions.

\ptspace
\partitle{Epic-Kitchens}
is a recently introduced large-scale dataset for cooking activities.
In total, it contains 274 videots performed by 28 actors in different kitchen setups.
Each video represents a cooking different cooking activity.
The average length of videos is 30 minutes, which makes it ideal for experimenting very long-range temporal modeling.
Originally, the task proposed by the dataset is classification on short video snippets, with average length of $\sim$3.7 seconds.
The provided labels are, therefore, the categories of objects, verbs and unit-actions in each video snippet.
However, the dataset does no provide video-level category.
That is why we consider all the object labels of a specific video as video-level label.
Hence, posing the problem as multi-label classification of these videos.
This setup is exactly the same used in Charades~\cite{sigurdsson2016hollywood} for video classification.
For performance evaluation, we use mean Average Precision (mAP), implemented in Sk-Learn~\cite{scikit-learn}.

\begin{table}[!ht]
\centering
\renewcommand{\arraystretch}{1.0}
\setlength\tabcolsep{2.3pt}
\begin{tabular}{lcc}
\specialrule{0.3mm}{.0em}{.3em}
Method           						                & Modality 		& mAP (\%) \\
\midrule
Two-stream~\cite{sigurdsson2017asynchronous}		    & RGB + Flow 	& 18.6 \\
Two-stream + LSTM~\cite{sigurdsson2017asynchronous}     & RGB + Flow 	& 17.8 \\
ActionVLAD~\cite{girdhar2017actionvlad}			        & RGB + iDT		& 21.0 \\
Temporal Fields~\cite{sigurdsson2017asynchronous}	    & RGB + Flow   	& 22.4 \\
		Temporal Relations~\cite{zhou2017temporal}	    & RGB		   	& 25.2 \\
\midrule
ResNet-152~\cite{charades2017algorithms}				& RGB        	& 22.8 \\
ResNet-152 + Timeception~\cite{hussein2018timeception}	& RGB        	& 31.6 \\
\midrule
I3D~\cite{carreira2017quo}								& RGB        	& 32.9 \\
I3D + ActionVLAD~\cite{girdhar2017actionvlad}	        & RGB        	& 35.4 \\
I3D + Timeception~\cite{hussein2018timeception}         & RGB        	& 37.2 \\
\textbf{I3D + VideoGraph}						        & RGB        	& \textbf{37.8} \\
\specialrule{0.3mm}{.0em}{.0em}
\end{tabular}
\caption{VideoGraph outperforms related works using the same backbone CNN.
Results are for Charades dataset.}
\label{tbl:4-2}
\vspace*{-10pt}
\end{table}

\begin{table*}[!ht]
\centering
\renewcommand{\arraystretch}{1.0}
\setlength\tabcolsep{8pt}
\begin{tabular}{lcccc}
\specialrule{0.3mm}{.0em}{.3em}
Method           						       & Breakfast Acc. (\%)  & Breakfast mAP (\%) & Epic-Kitchens mAP (\%) \\
\midrule
ResNet-152~\cite{charades2017algorithms}		        & 41.13     &  32.65     & --   \\
ResNet-152 + ActionVLAD~\cite{girdhar2017actionvlad}    & 55.49     &  47.12     & --   \\
ResNet-152 + Timeception~\cite{hussein2018timeception}  & 57.75     &  48.47     & --   \\
\textbf{ResNet-152 + VideoGraph}				        & \textbf{59.12} & \textbf{49.38}  & --   \\
\midrule
I3D~\cite{carreira2017quo}						        & 58.61   & 47.05  & 48.86     \\
I3D + ActionVLAD~\cite{girdhar2017actionvlad}	        & 65.48   & 60.20  & 51.45    \\
I3D + Timeception~\cite{hussein2018timeception}		    & 67.07   & 61.82  & \textbf{55.46}    \\
\textbf{I3D + VideoGraph}						        & \textbf{69.45}   & \textbf{63.14} & 55.32 \\
\specialrule{0.3mm}{.0em}{.0em}
\end{tabular}
\caption{VideoGraph outperforms related works using the same backbone CNN. We experiment 2 different backbones: I3D and ResNet-152. We experiment on two different tasks of Breakfast: single-label classification of activities and multi-label classification of unit-actions. And for Epic-Kitchens, we experiment on the multi-label classification.}
\label{tbl:4-1}
\vspace*{-5pt}
\end{table*}

\ptspace
\partitle{Charades}
is a dataset for multi-label classification of action videos.
It consists of 8k, 1.2k and 2k video for training, validation and testing, respectively.
is multi-label, action classification, video dataset with 157 classes.
Each video spans 30 seconds and comprises of 6 unit-actions, on average.
This is why we choose Charades, as it fits perfectly to the needs of this paper.
For evaluation, we use mAP, as detailed in~\cite{sigurdsson2016hollywood}.

\subsection{Experiments on Benchmarks}
In this section, we experiment and evaluate VideoGraph on benchmark datasets: Breakfast, Charades and Epic-Kitchens, and we compare against related works.
We choose two strong methods to compare against.
The first is Timeception~\cite{hussein2018timeception}.
The reason is that it can model 1k timesteps, which is up to a minute-long video.
Another reason is that Timeception is an order-ware temporal method.
The second related work is ActionVLAD~\cite{girdhar2017actionvlad}.
The reason is that it is a strong example of orderless method.
It also can aggregate temporal signal for very long videos.

VideoGraph resides on top of backbone CNN, be it spatial 2D CNN, or spatio-temporal 3D CNN.
So, in our comparison, we use two backbone CNNs, namely ResNet-152~\cite{he2016deep} and I3D~\cite{carreira2017quo}.
By default, I3D is designed to model a short video segment of 8 frames.
But thanks to the fully-convolutional architecture, I3D can indeed process minutes-long video.
This is made possible by average pooling the features of many videos snippets, in logit layer, \ie before softmax activation~\cite{carreira2017quo}.
ResNet-152 is a frame-level classifier. To extend it to video classification, we follow the same approach used in I3D and average pool the logits, \ie before softmax.
In all the following comparisons, we use 512 frames, or 64 segments, per video as input to I3D.
And we use 64 frames per video as and input to ResNet-152.

\ptspace
\partitle{Breakfast.}
Each video in this dataset depicts a complex breakfast activity.
Thus, the task inhand is single-label classification.
The evaluation metric used is the classification accuracy.
We experiment our model on Breakfast, and we compare against baseline methods.
The results are reported in table~\ref{tbl:4-1}.

\ptspace
\partitle{Epic-Kitchens.}

When comparing VideoGraph against related works, see table~\ref{tbl:4-1}, Timeception and VideoGraph, we notice that we are on bar with Timeception.
VideoGraph performs better when trained on single-label video dataset, where each video has one label.
This gives VideoGraph an ample opportunity to tailor the graph-inspired representation for each class.
However, as mentioned, we pose the task in Epic-Kitchen as multi-label classification.
That is, no single category for a video.
That's when VideoGraph does not perform as good.

\ptspace
\partitle{Charades.}
In this experiment, we evaluate our model on Charades dataset. And we compare the performance against recent works. The results are reported in Table~\ref{tbl:4-2}. VideoGraph improves the performance of the backbone CNN. For VideoGraph, Charades is particularly challenging dataset, for two reasons.
First, the average video length is 30 seconds, and VideoGraph learns better representstion for long-range videos.
Second, it is a multi-label classification, and that's when VideoGraph is not able to learn category-specific unique graph.

\begin{figure}
\begin{subfigure}[t]{0.22\textwidth}
\includegraphics[width=\textwidth]{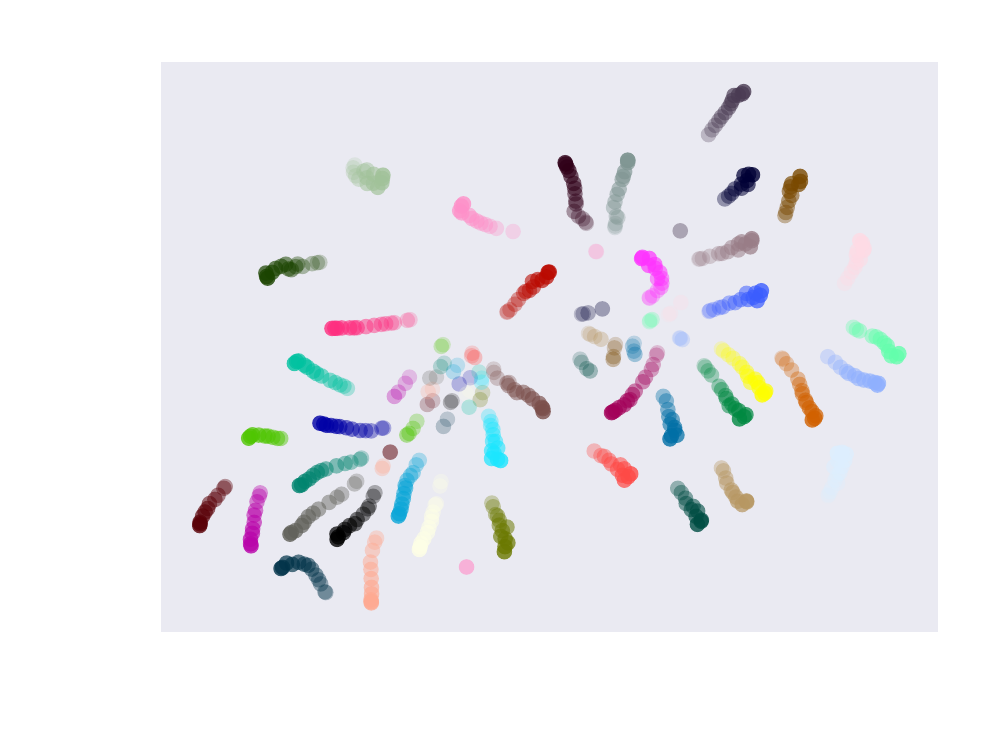}
\caption{}
\vspace*{-2mm}
\label{fig:4-1-1}
\end{subfigure}\hfill
\begin{subfigure}[t]{0.22\textwidth}
\includegraphics[width=\textwidth]{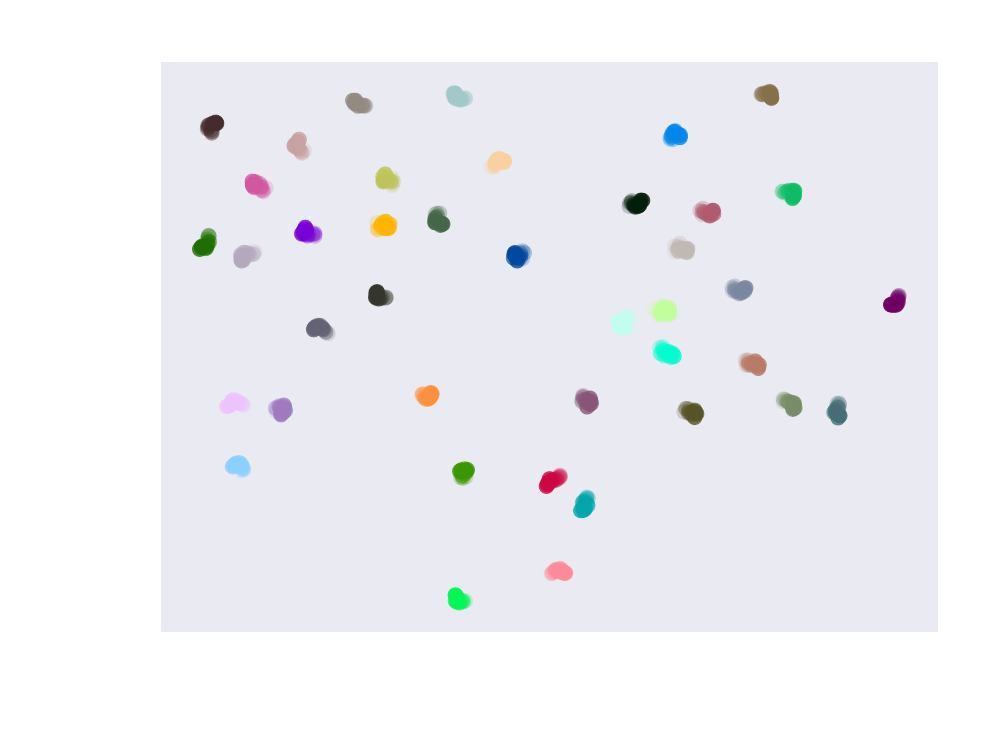}
\caption{}
\vspace*{-2mm}
\label{fig:4-1-2}
\end{subfigure}
\caption{Visualization of the learned graph nodes.
In the first 20 epoch during training (left), VideoGraph updates the node features $\hat{Y}$ to increase the pairwise distance between them.
That is, VideoGraph learns discriminant representations of the nodes.
In the last 20 epoch during training (right), the learning cools down and barely their representation is updated.
We visualize using t-SNE~\cite{maaten2008visualizing}.}
\vspace*{-5mm}
\label{fig:4-1}
\end{figure}

\begin{figure}[!ht]
\begin{center}
\includegraphics[trim=3mm 0mm 0mm 0mm, width=0.99\linewidth]{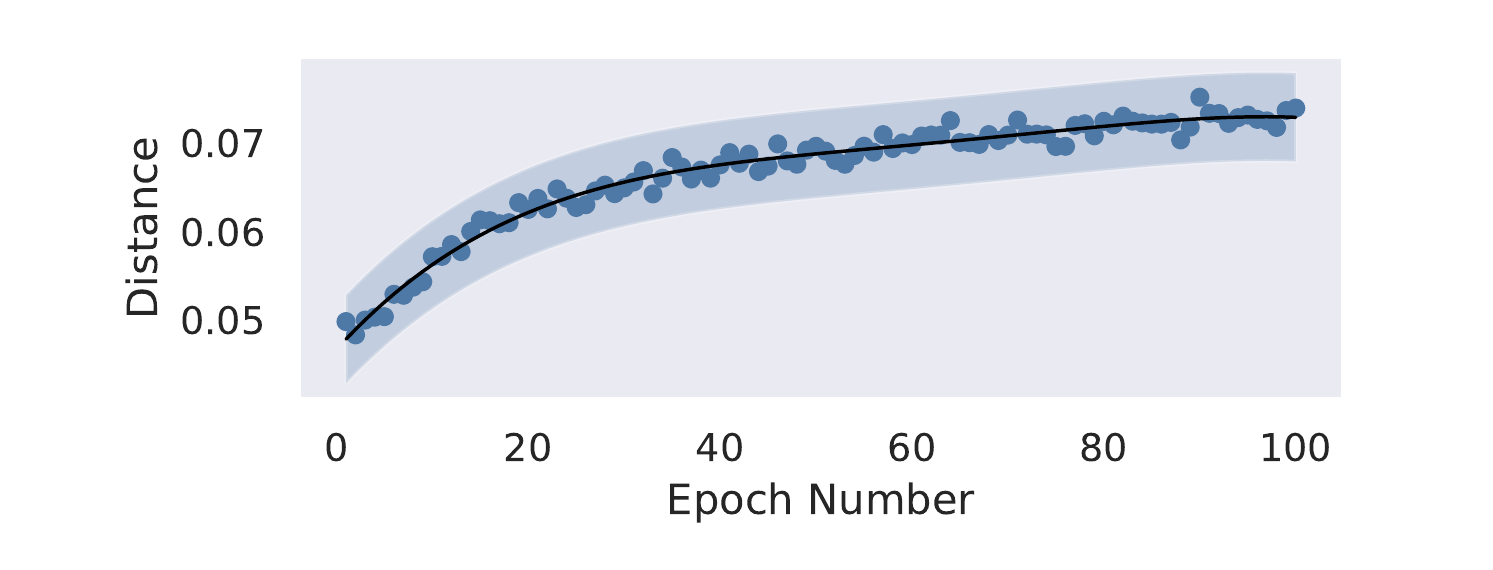}
\end{center}
\vspace*{-5mm}
\caption{
The pairwise Euclidean distances between normalized latent concepts $\hat{Y}$ increases rapidly in the beginning of the training, but it converges in the end.}
\label{fig:4-4}
\vspace*{-5mm}
\end{figure}

\begin{figure*}
\begin{subfigure}[t]{0.195\textwidth}
\includegraphics[trim=0mm 0mm 0mm 0mm, width=\textwidth]{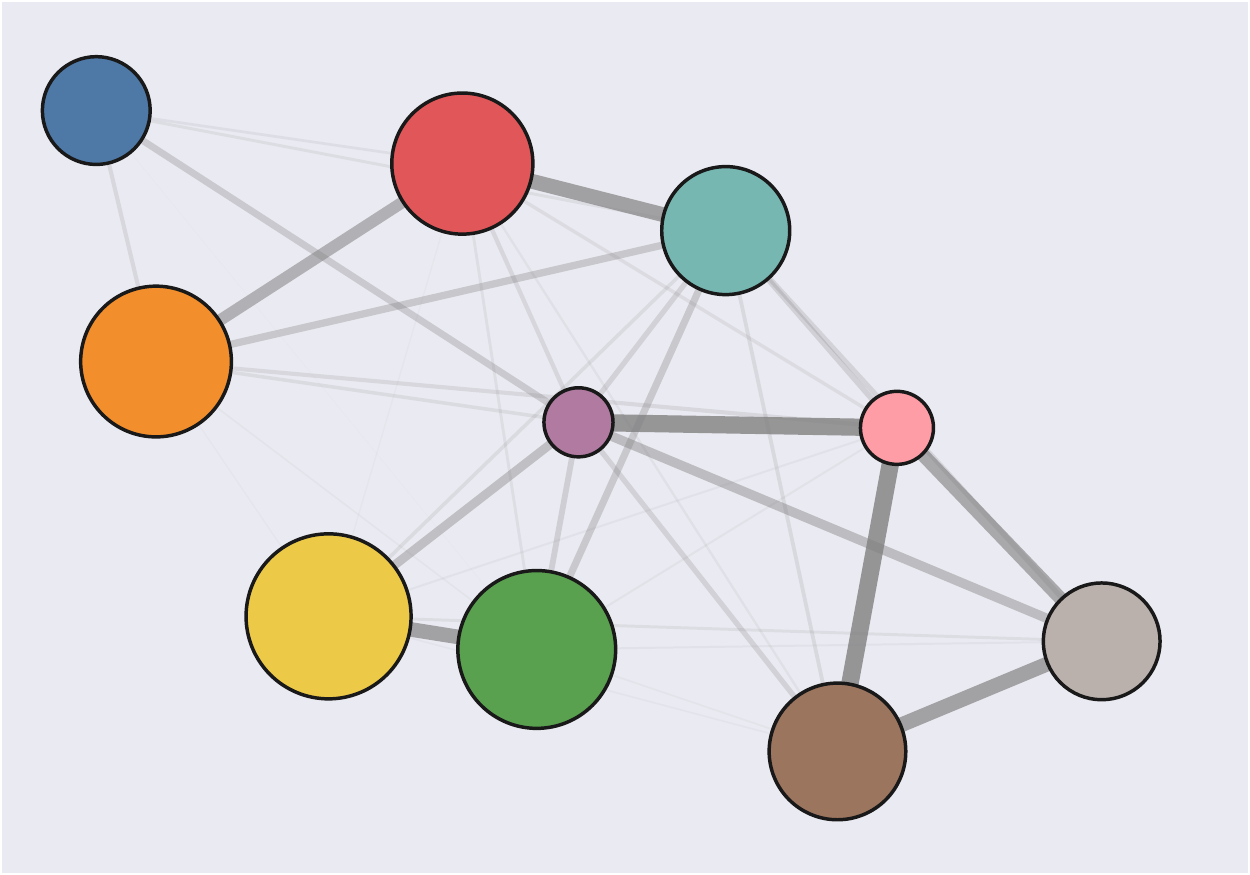}
\caption{Making Cereals}
\vspace*{2mm}
\label{fig:4-5-1}
\end{subfigure}
\hfill
\begin{subfigure}[t]{0.195\textwidth}
\includegraphics[trim=0mm 0mm 0mm 0mm, width=\textwidth]{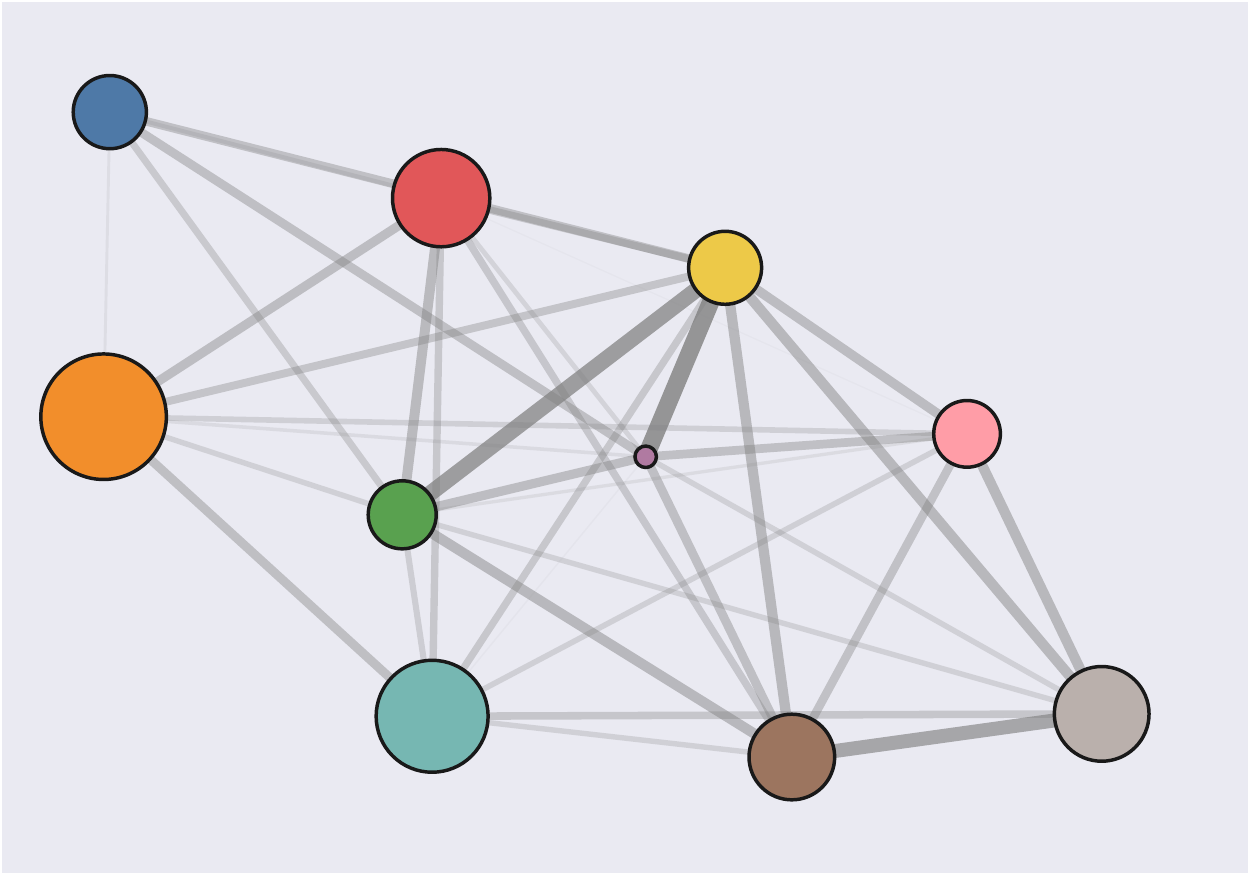}
\caption{Preparing Coffee}
\vspace*{0mm}
\label{fig:4-5-2}
\end{subfigure}
\hfill
\begin{subfigure}[t]{0.195\textwidth}
\includegraphics[trim=0mm 0mm 0mm 0mm, width=\textwidth]{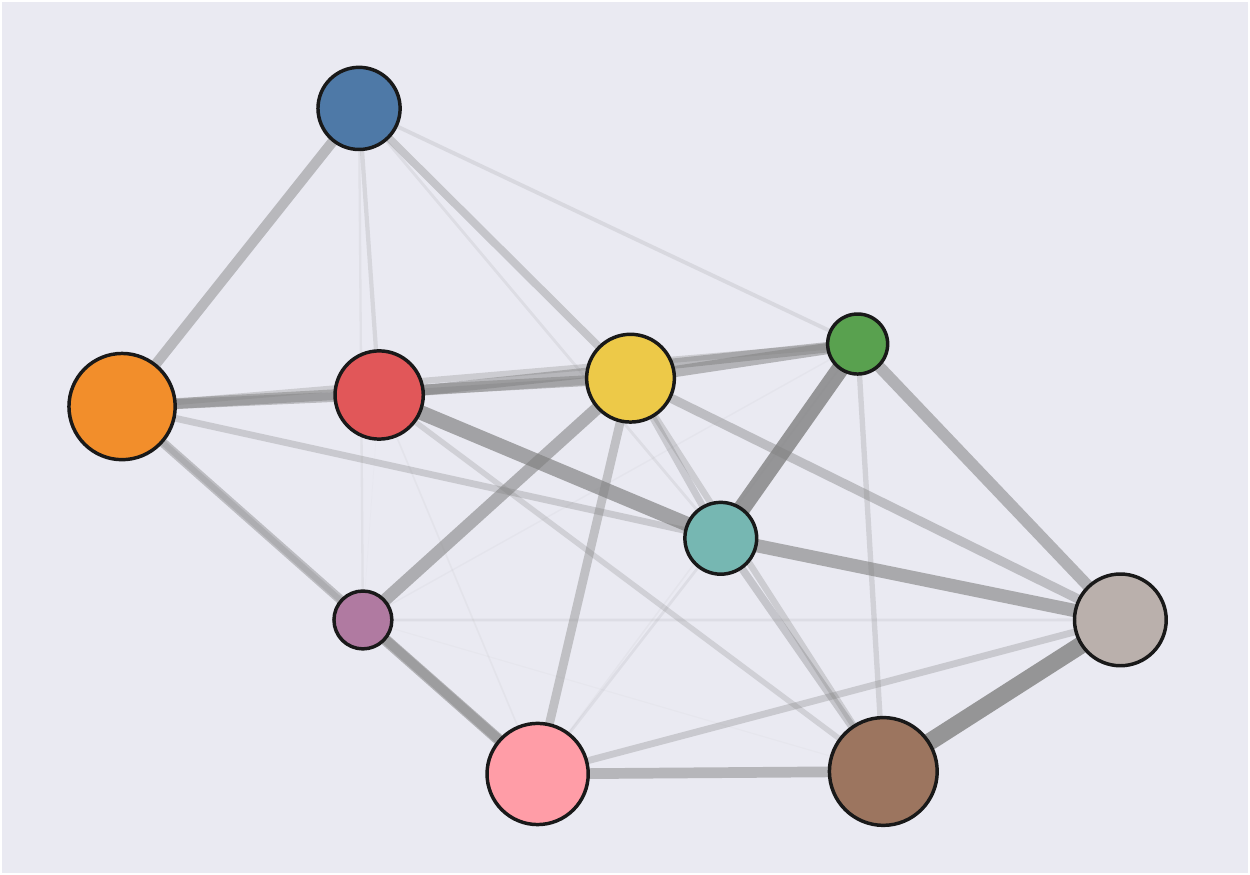}
\caption{Frying Eggs}
\vspace*{0mm}
\label{fig:4-5-3}
\end{subfigure}
\hfill
\begin{subfigure}[t]{0.195\textwidth}
\includegraphics[trim=0mm 0mm 0mm 0mm, width=\textwidth]{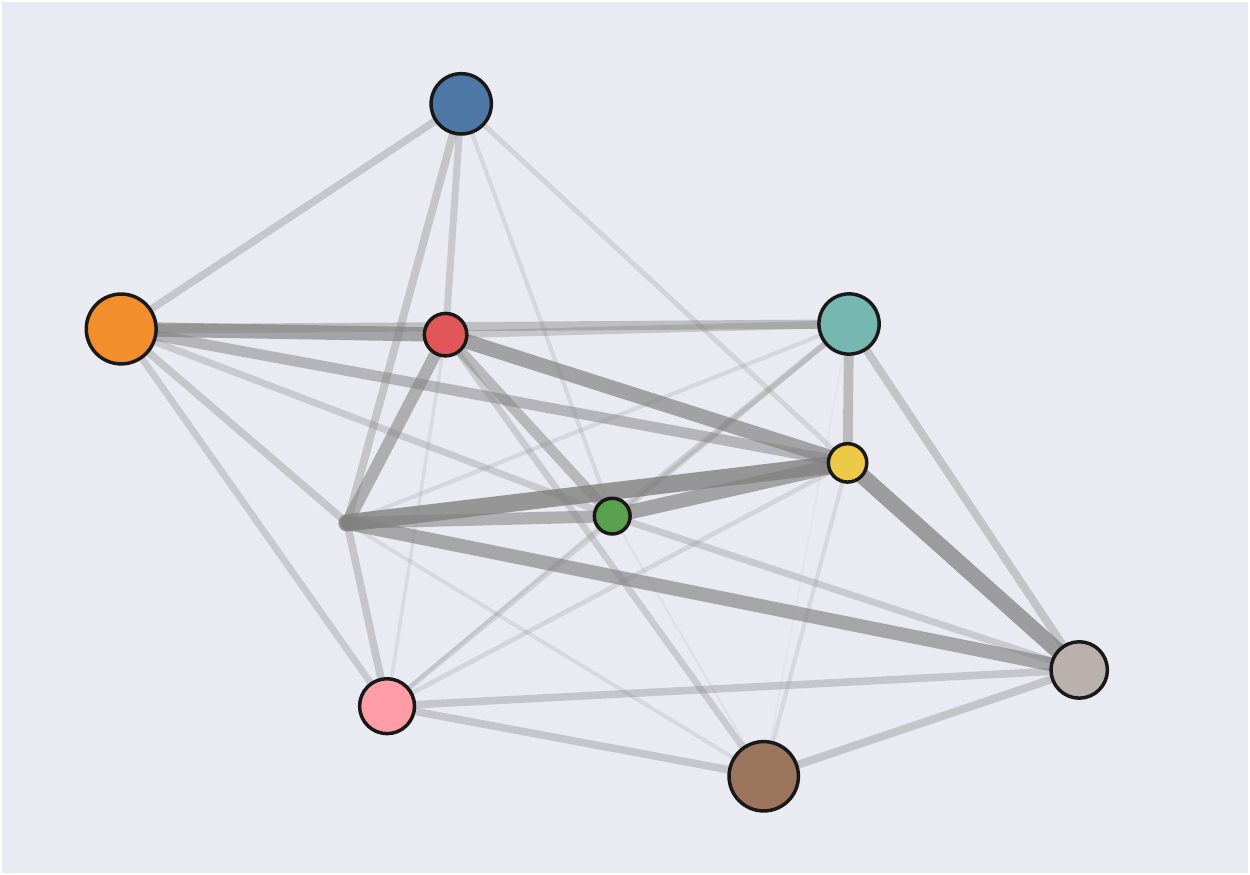}
\caption{Making Juice}
\vspace*{0mm}
\label{fig:4-5-4}
\end{subfigure}
\hfill
\begin{subfigure}[t]{0.195\textwidth}
\includegraphics[trim=0mm 0mm 0mm 0mm, width=\textwidth]{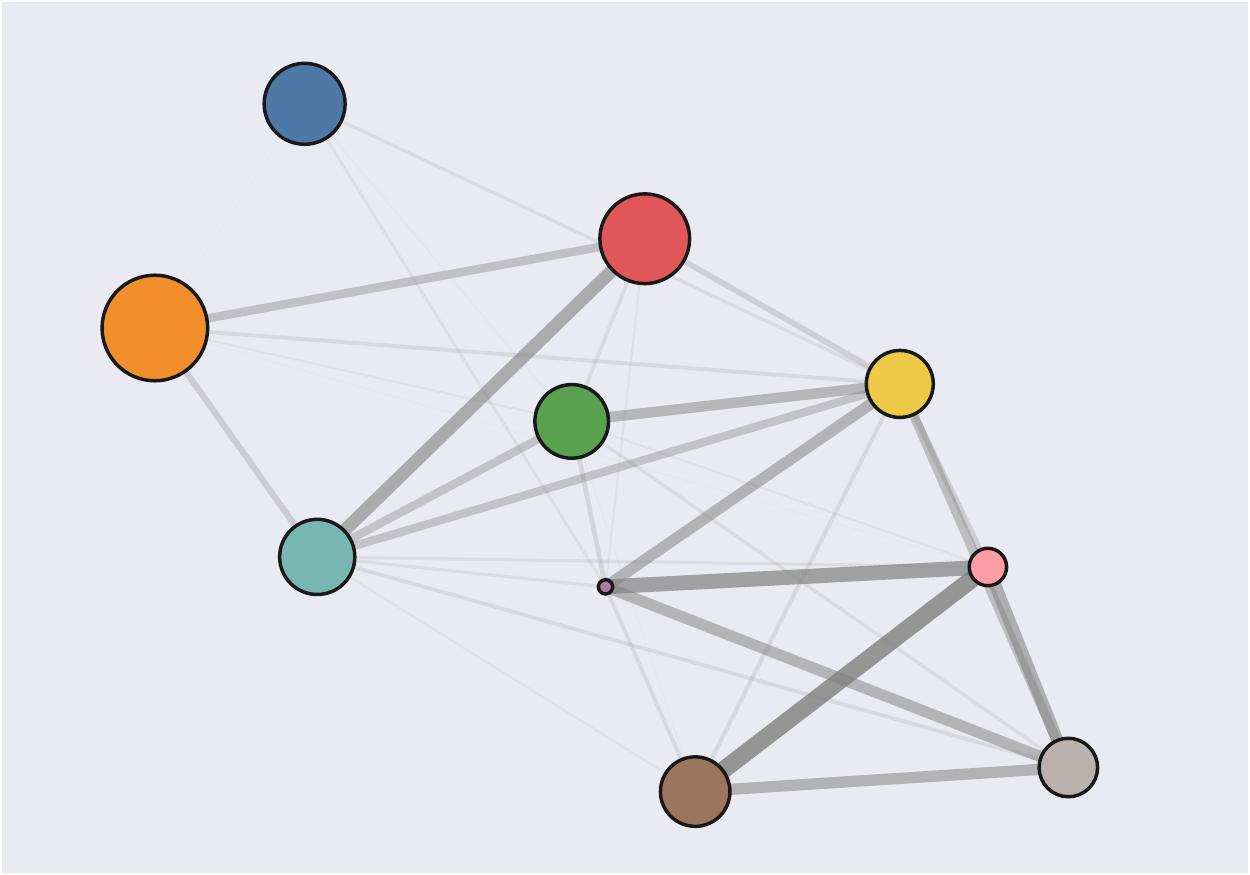}
\caption{Preparing Milk}
\vspace*{0mm}
\label{fig:4-5-5}
\end{subfigure}
\vfill
\begin{subfigure}[t]{0.195\textwidth}
\includegraphics[trim=0mm 0mm 0mm 0mm, width=\textwidth]{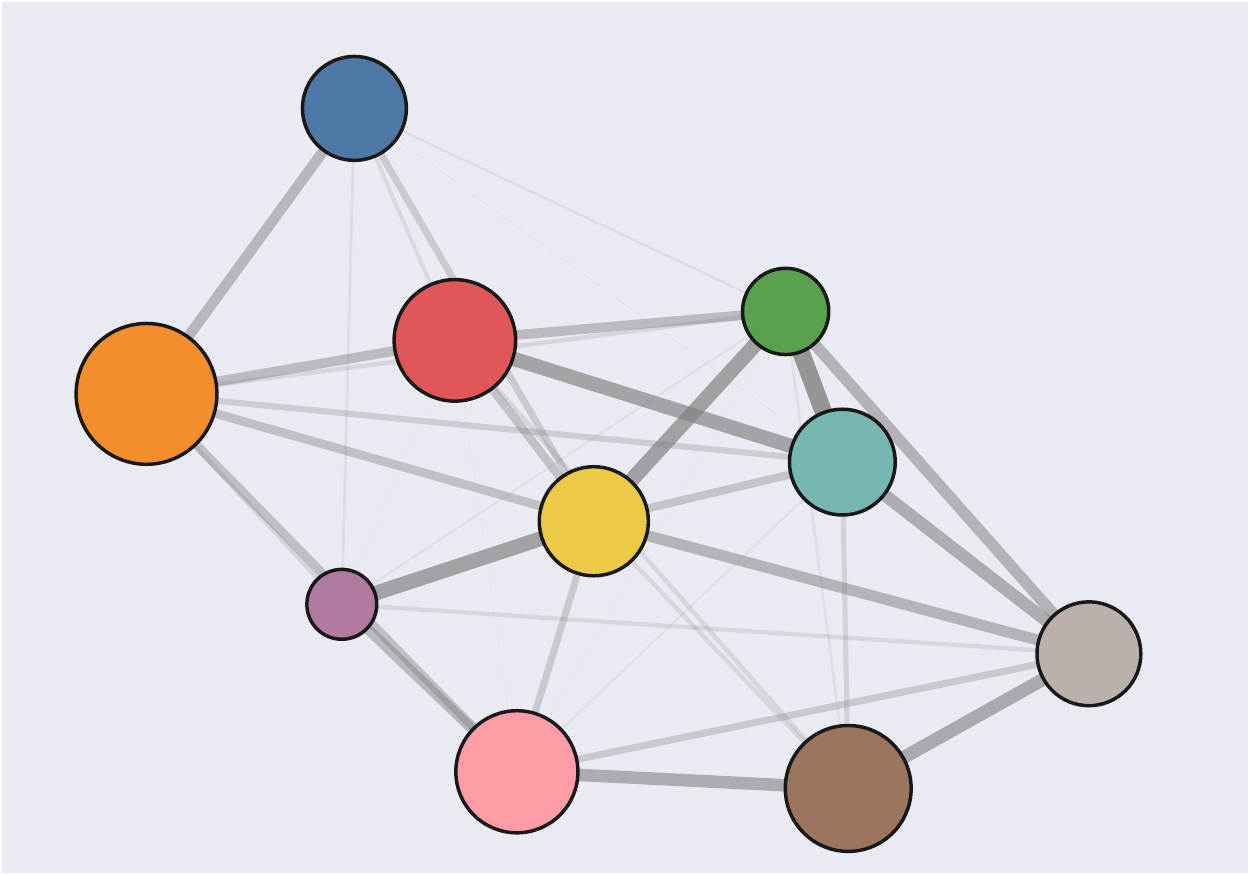}
\caption{Making Pancake}
\vspace*{0mm}
\label{fig:4-5-6}
\end{subfigure}
\hfill
\begin{subfigure}[t]{0.195\textwidth}
\includegraphics[trim=0mm 0mm 0mm 0mm, width=\textwidth]{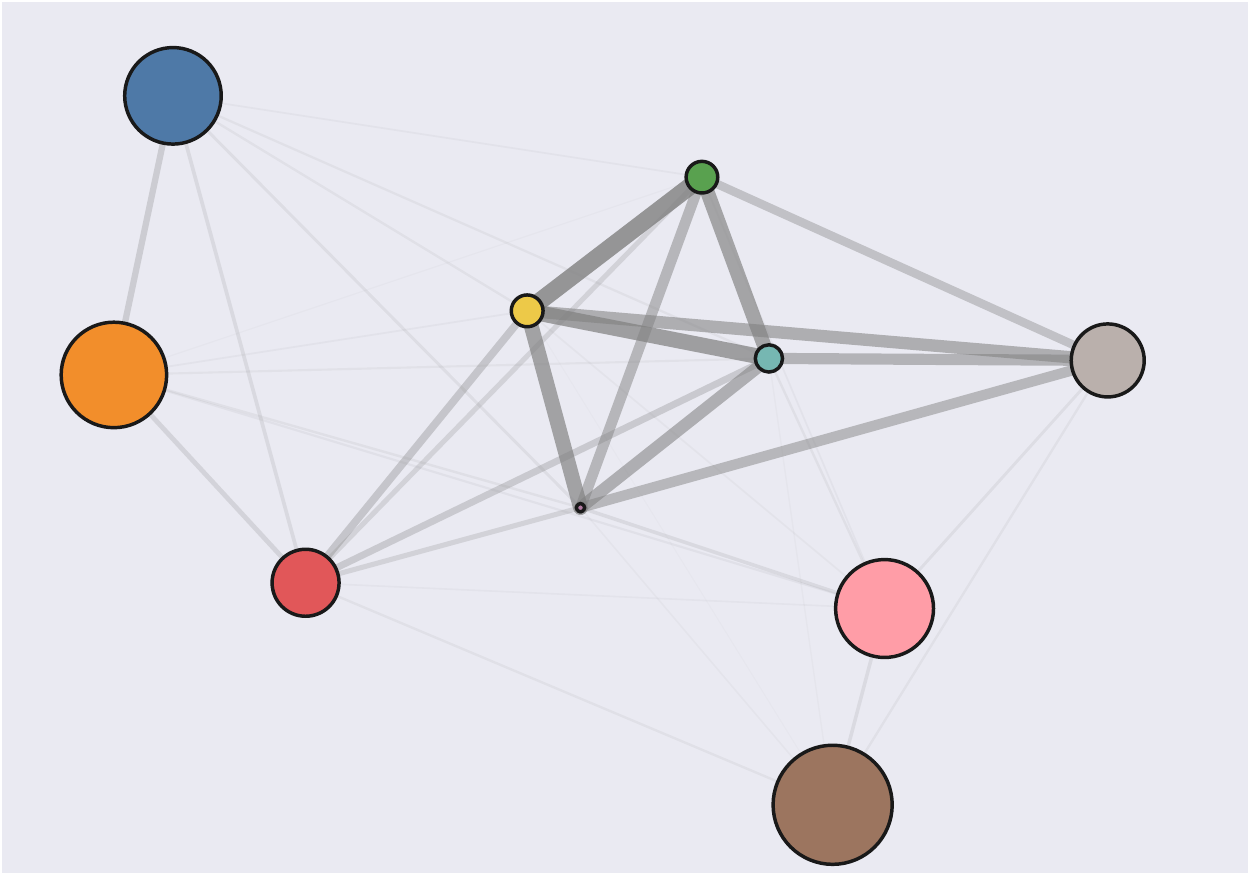}
\caption{Making Salad}
\vspace*{0mm}
\label{fig:4-5-7}
\end{subfigure}
\hfill
\begin{subfigure}[t]{0.195\textwidth}
\includegraphics[trim=0mm 0mm 0mm 0mm, width=\textwidth]{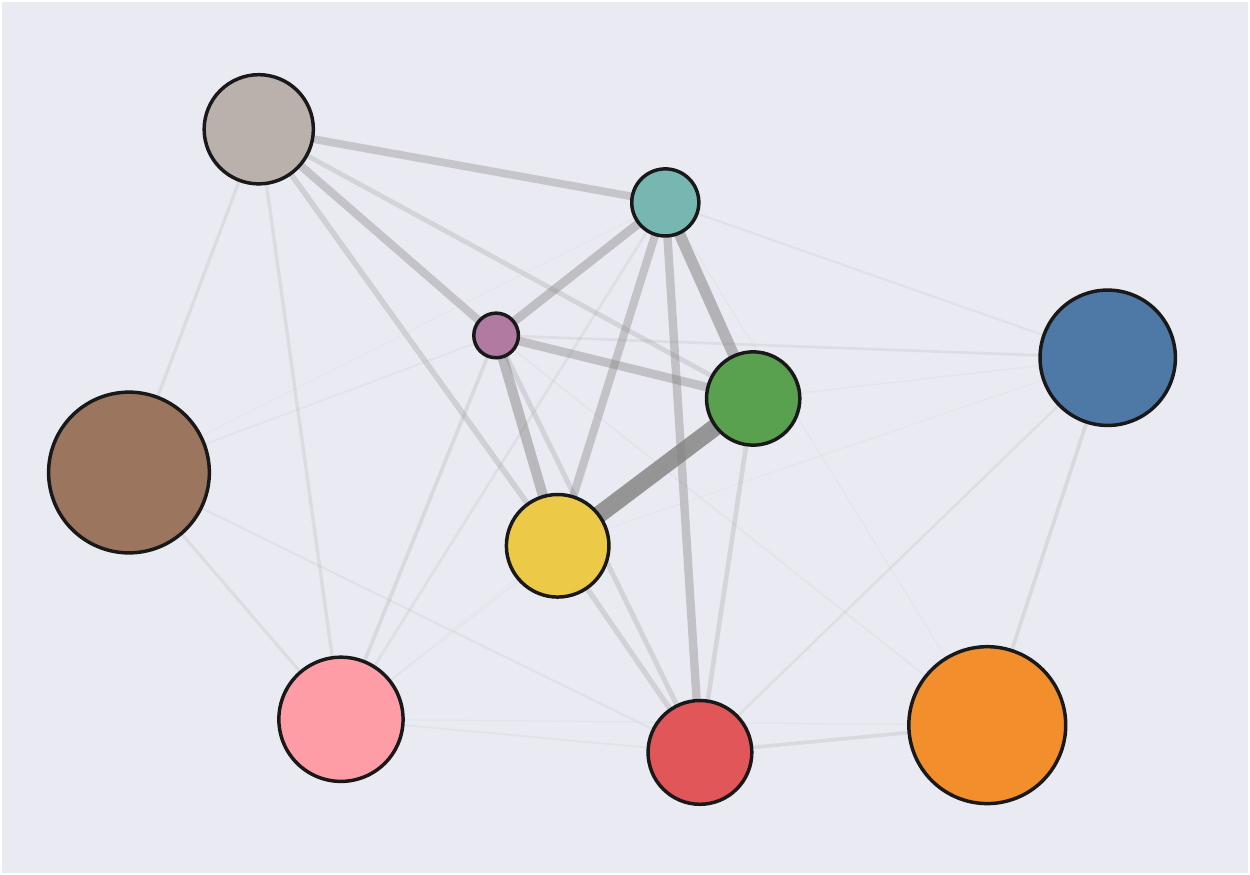}
\caption{Making Sandwich}
\vspace*{0mm}
\label{fig:4-5-8}
\end{subfigure}
\hfill
\begin{subfigure}[t]{0.195\textwidth}
\includegraphics[trim=0mm 0mm 0mm 0mm, width=\textwidth]{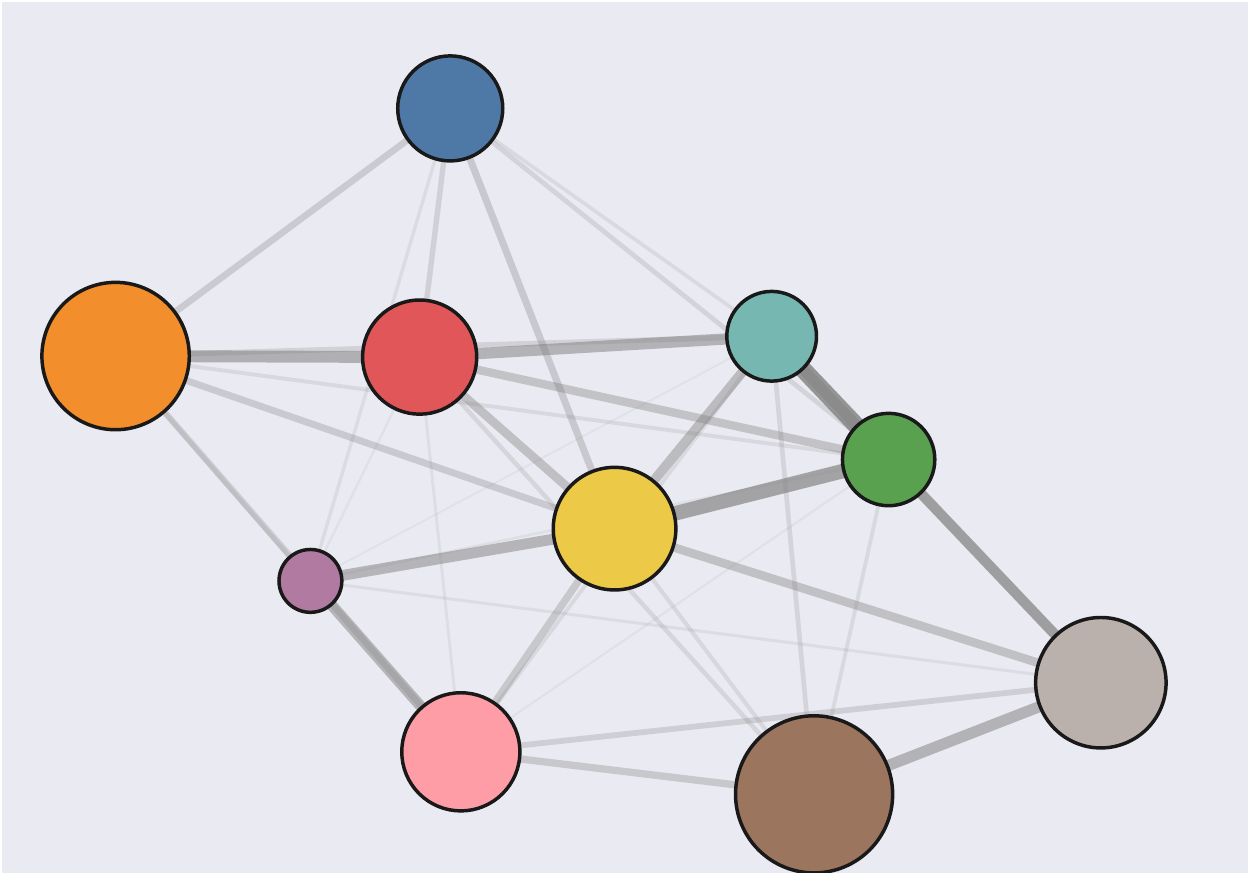}
\caption{Making Scrambled Egg}
\vspace*{0mm}
\label{fig:4-5-9}
\end{subfigure}
\hfill
\begin{subfigure}[t]{0.195\textwidth}
\includegraphics[trim=0mm 0mm 0mm 0mm, width=\textwidth]{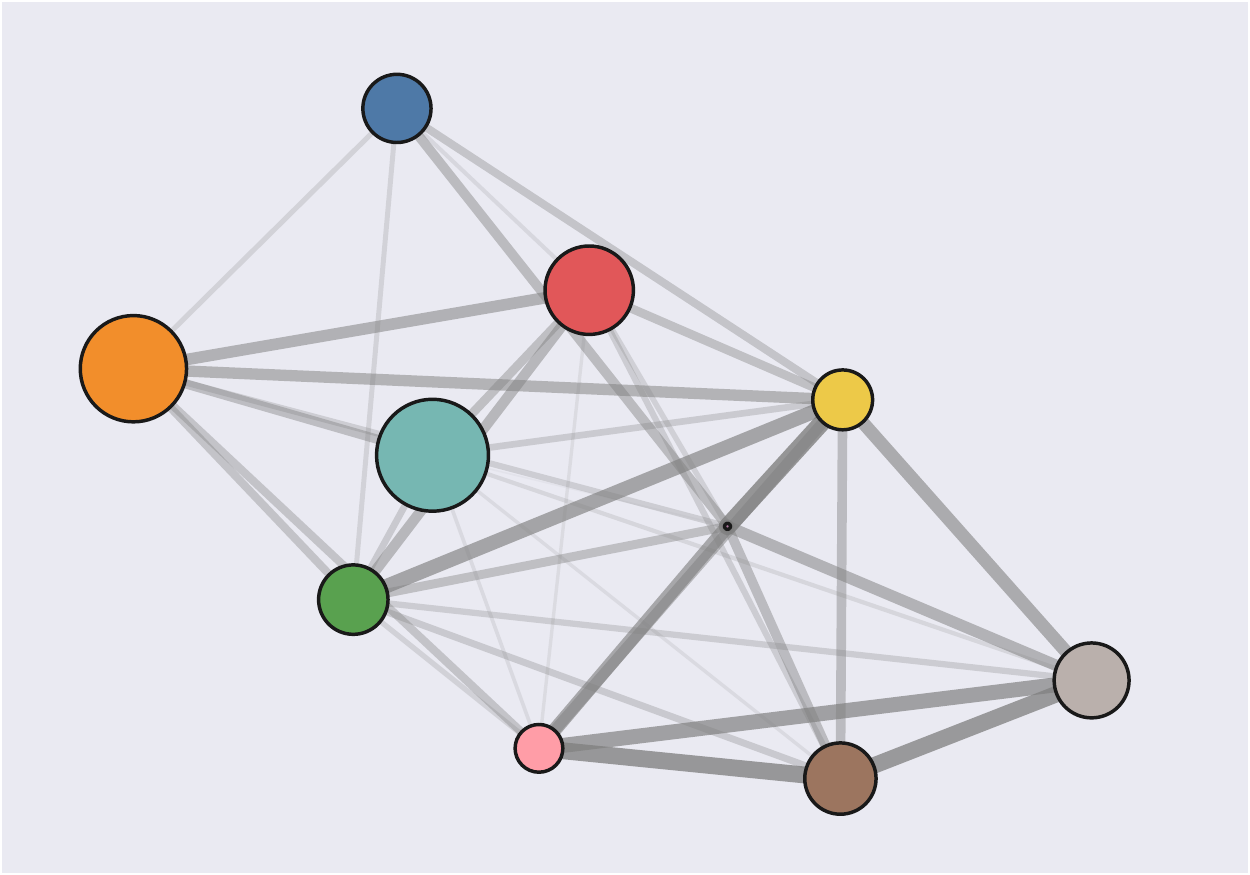}
\caption{Preparing Tea}
\vspace*{0mm}
\label{fig:4-5-10}
\end{subfigure}
\vfill
\begin{subfigure}[t]{1.0\textwidth}
\includegraphics[trim=2mm 5mm 2mm -8mm, width=\textwidth]{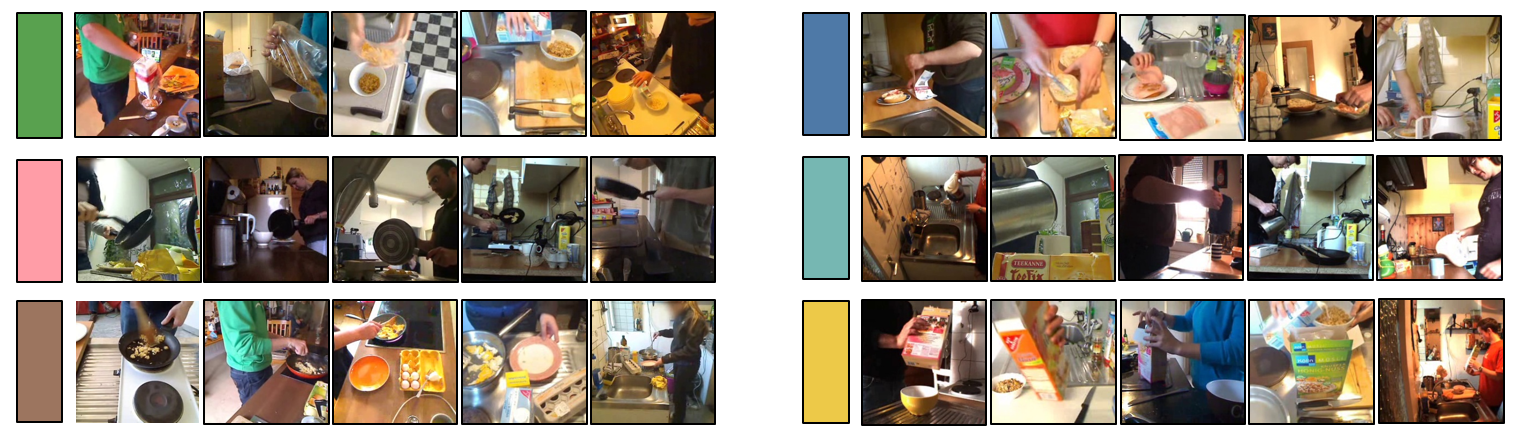}
\caption{Top related images to the nodes.
These nodes are related to: \textcolor{colornode05}{\newmoon} cereal, \textcolor{colornode08}{\newmoon} pan, \textcolor{colornode09}{\newmoon} eggs, \textcolor{colornode01}{\newmoon} sandwich, \textcolor{colornode04}{\newmoon} kettle, and \textcolor{colornode06}{\newmoon} foodbox.}
\vspace*{0mm}
\label{fig:4-5-11}
\end{subfigure}
\caption{We visualize the relationship discovered by the first layer of graph embedding.
Each sub-figure is related to one of the 10 activities in Breafast dataset.
In each graph, the nodes represent the latent concepts learned by graph-attention block. Node size reflects how dominant the concept, while graph edges emphasize the relationship between these nodes.}
\label{fig:4-5}
\end{figure*}

\subsection{Learned Graph Nodes}
The proposed node attention block, see figure~\ref{fig:3-3}{\color{red}a}, learns latent concept representation $\hat{Y}$ using fully-connected layer.
This learning is conditioned on the initial value $Y$.
We found that this initial value is crucial for VideoGraph to converge.
We experiment with 3 different types of initialization: \textit{i.} random values, \textit{ii.} Sobol sequence and \textit{iii.} k-means centroids.
Random values seems to be a natural choice, as all the learned weights in the model are randomly initialized before training.
Sobol sequence is a plausible choice, as the sequence guarantees low discrepancies between the initial values.
The last choice has proven to be successful in ActionVLAD~\cite{girdhar2017actionvlad}. The centroids are obtained by clustering the feature maps of the last convolutional layer of the backbone CNN.
However, we do not find one winning strategy across the benchmarks used.
We find that Sobol sequence is the best choice for training on Epic-Kitchens and Charades.
While the random initialization gives the best results on Breakfast.
In table~\ref{tbl:4-2}, we report the performance of VideoGraph using different initialization choices for the latent concepts $Y$.
In all cases, we see in figure~\ref{fig:4-1} that the node attention layer successfully learns discriminant representations of latent concepts, as the training proceedes.
In other words, the networks learns to increase the Euclidean distance between each pair of latent concepts.
This is further demonstrated in figure~\ref{fig:4-4}.

\begin{table}[!ht]
\centering
\renewcommand{\arraystretch}{1.0}
\setlength\tabcolsep{3pt}
\begin{tabular}{lcc}
\specialrule{0.3mm}{.0em}{.3em}
Initialization      			& Epic-Kitchen mAP         & Breakfast Acc. \\
\midrule
Random		                    & 54.12        &            \textbf{69.45}  \\
Sobol                           & \textbf{55.46}            &      65.61    \\
K-means Centroids		        & 52.47                 &      ---          \\
\specialrule{0.3mm}{.0em}{.0em}
\end{tabular}
\caption{The initialization of the latent concepts is crucial for learning better representation $\hat{Y}$.
We experimented with 3 choices: random, sobol, and k-mean clustering.
Yet, there seems not to be one winning choice across different datasets.}
\label{tbl:4-3}
\vspace*{-10pt}
\end{table}

\subsection{Learned Graph Edges}
There are two types of graph edges, \textit{i.e.} relationships, uncovered by VideoGraph.
First, the timewise edges, \ie how the nodes transition over time.
Second, the nodewise edges, \ie relationships between nodes themselves.
To this end, we depend on the activation output of the second graph embedding layer.
In other words, we extract the ReLU activation values. For $M$ videos belonging to a specific human activity, the activation values are $z_1 \in \mathbb{R}^{M \timesnarrow N \timesnarrow T \timesnarrow C}$, where $C$ is the number of channels, $T$ is the number of timesteps, and $N$ is the number of nodes.
First, we average the activations for all the videos, resulting in $z_2 \in \mathbb{R}^{N \timesnarrow T \timesnarrow C}$.
Then, we average pool the activations over the temporal dimension, so we have $z_3 \in \mathbb{R}^{N \timesnarrow C}$, summarizing the nodes representations for all the videos belonging to the specific activity.
Finally, we measure the pairwise Euclidean distance between each pair in $z_3$.
To plot the graph depicting the activity, we use these distances as the edge between the nodes.
And to plot the nodes, we sum up the activations over the channel dimension in $z_3$.
The result $z_4 \in \mathbb{R}^{N}$ is a scalar value reflecting the importance of the node to the activity.
The graph is plotted using Fruchterman-Reingold force-directed algorithm, implemented in~\cite{networkx}.
Figure~\ref{fig:4-5} shows 10 different graph, each belonging to one human activity.

\ptspace
\partitle{Importance of Temporal Structure.}
In this experiment, we validate by how much VideoGraph depends on the temporal structure and weak temporal order to recognize the human activities.
To this end, we choose Breakfast, as it is temporally well-structured dataset.
VideoGraph is trained on ordered set of $64$ timesteps.
We alter the temporal order of these timesteps and test the performance of VideoGraph.
We use different alterations: \textit{i.} random order, and \textit{ii.} reversed order.
Then, we measure the performance of VideoGraph, as well as baselines, on Breakfast testset.

\begin{table}[!ht]
\centering
\renewcommand{\arraystretch}{1.0}
\setlength\tabcolsep{6pt}
\begin{tabular}{lccc}
\specialrule{0.3mm}{.0em}{.3em}
Temporal Structure     	& Reversed ($\downarrow$\%)  & Random ($\downarrow$\%) \\
\midrule
I3D              & 0.0 & 0.0   \\
I3D + ActionVLAD & 0.0 & 0.0   \\
\midrule
I3D + Timeception    &   44.1  & 56.2    \\
I3D + VideoGraph     &   22.5  & 55.9   \\
\specialrule{0.3mm}{.0em}{.0em}
\end{tabular}
\caption{The drop of performance of VideoGraph and other models when changing the temporal order of the input video.
Both VideoGraph and Timeceptions suffer huge drop in performace, as both are order-aware methods.
On the other hand, ActionVLAD retains the same performance, as it is orderless method.}
\label{tbl:4-4}
\vspace*{-10pt}
\end{table}

\begin{figure}[!ht]
\begin{center}
\includegraphics[trim=8mm 0mm 0mm 0mm, width=1.0\linewidth]{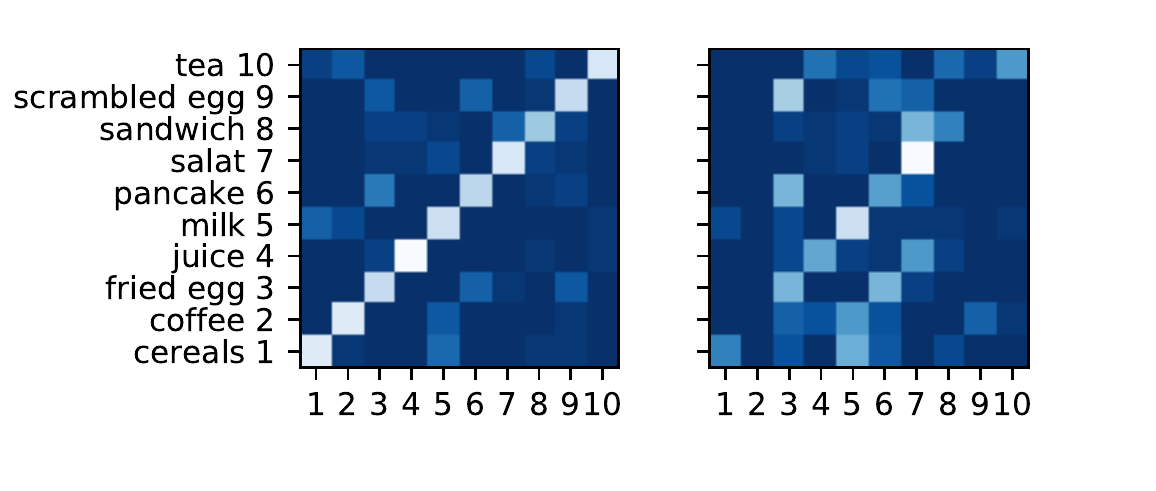}
\end{center}
\vspace*{-5mm}
\caption{Confusion matrix for recognizing the 10 activity of Breakfast.
VideoGraph is trained on random (right) \textit{v.s.} correct temporal order (left).
It mistakes ``scrambled egg" for ``fried egg" if temporal order is neglected.}
\label{fig:4-11}
\vspace*{-5mm}
\end{figure}

We notice, from table~\ref{tbl:4-4}, a huge drop in performance for both VideoGraph and Timeception.
However, as expected, no drop in performance for ActionVLAD, as it is completely orderless model.
The conclusion is VideoGraph encodes the temporal structure of the human activities in breakfast.
Added to this, it suffered slightly less drop in performance than Timeception. More importantly, figure~\ref{fig:4-11} shows the confusion matrix of classifiyng the videos of Breakfast using two cases: \textit{i} natural order of temporal video segments, and \textit{ii.} random order of the video segments.
We notice video graph makes more mistakes when trained on random order.
It mistakes ``scrambled egg" for ``fried egg" if temporal order is neglected.

\section{Conclusion}\label{sec:conclusions}
To successfully recognize minutes-long human activities such as ``preparing breakfast'' or ``cleaning the house'', we argued that a successful solution needs to capture both the whole picture and attention to details. To this end, we proposed VideoGraph, a graph-inspired representation to model the temporal structure of such long-range human activities.
Firstly, thanks to the node attention layer, VideoGraph can learn the graph nodes. This alleviate the need of node-level annotation, which is prohibitive and expensive in nowadays video dataset.
Secondly, we proposed graph embedding layer.
It learns the relationship between graph nodes and how these nodes transition over time.
Also, it compresses the graph representation to be feed for a classifier.
We demonstrated the effectiveness of VideoGraph on three benchmarks: Breakfast, Epic-Kitchens and Charades.
VideoGraph achieves good performance on the three of them.
We also discussed some of the upsides and downside of VideoGraph.

{\small
\bibliographystyle{unsrt}
\bibliography{main}
}

\end{document}